# NN-Copula-CD: A Copula-Guided Interpretable Neural Network for Change Detection in Heterogeneous Remote Sensing Images

Weiming Li, Xueqian Wang*, Gang Li, Baocheng Geng, and Pramod K. Varshney

## Abstract

Change detection (CD) in heterogeneous remote sensing images has been widely used for disaster monitoring and land-use management. In the past decade, the heterogeneous CD problem has significantly benefited from the development of deep neural networks (DNNs). However, the purely data-driven DNNs perform like a black box where the lack of interpretability limits the trustworthiness and controllability of DNNs in most practical CD applications. As a powerful knowledge-driven tool, copula theory performs well in modeling relationships among random variables. To enhance the interpretability of existing neural networks for CD, we propose a knowledge-data-driven heterogeneous CD method based on a copula-guided neural network, named NN-Copula-CD. In our NN-Copula-CD, the mathematical characteristics of copula are employed as the loss functions to supervise a neural network to learn the dependence between bi-temporal heterogeneous superpixel pairs, and then the changed regions are identified via binary classification based on the degrees of dependence of all the superpixel pairs in the bi-temporal images. We conduct in-depth experiments on three datasets with heterogeneous images, where both quantitative and visual results demonstrate the effectiveness of our proposed NN-Copula-CD method.

W. Li, X. Wang, and G. Li are with the Department of Electronic Engineering, Tsinghua University, Beijing 100084, China.
B. Geng is with the Department of Computer Science, The University of Alabama at Birmingham, Birmingham, AL 35294 USA.
P. K. Varshney is with the Department of Electrical Engineering and Computer Science, Syracuse University, Syracuse, NY 13244 USA.

* Corresponding author: Xueqian Wang (Email: wangxueqian@mail.tsinghua.edu.cn)

# I. Introduction

Change detection (CD) is an important technology that has been applied to a large number of earth observation tasks [2, 3], such as disaster monitoring [4, 5], land-use management [6, 7], and ecosystem protection [8]. The aim of CD is to monitor natural and man-made changes by comparing the remote sensing images of the same geographical location acquired at different times [9]. Depending on the modalities of remote sensing images involved in CD, the CD tasks can be categorized into homogenous and heterogeneous CD tasks. The homogeneous CD task relies on remote sensing images of the same modality, and has been extensively investigated during the past decades [10]. However, the satellites' revisit cycles and weather conditions often limit the timeliness of homogenous CD in many practical situations, especially for urgent events [11, 12]. With the proliferation of satellites equipped with various imaging sensors in recent years, heterogeneous remote sensing images are becoming increasingly accessible and offer more timely information on land-cover objects [11, 13]. The heterogeneous CD task has demonstrated its great potential and utility in many CD applications, especially for time-sensitive emergencies like disaster monitoring, and has been attracting more and more attention from scholars and practitioners [11, 12].

Due to the heterogeneity in heterogeneous remote sensing images, the direct pixel comparison using algebraic calculation [14-16] for the CD task is not feasible. To solve this problem, one intuitive idea [17] is to classify the heterogeneous images separately with a set of categories, and then perform decision-level fusion based on the classification results to determine the location of changes. The post-classification treatment in [17] for heterogeneous images suppresses the false alarms/missed detections caused by the impact of heterogeneity to a certain extent, yet its performance relies heavily on the precision of the classification algorithms employed for each image.

Another intuitive idea for the heterogeneous CD issue is to transform the bi-temporal images into the same space where they are easier to be compared [13]. Following this idea, earlier works [18] utilized the 3-D information of the land-cover objects and imaging parameters of satellites to predict the characteristics of pre-change land-cover objects in the post-change image feature space (assuming no change event occurred). Then, the similarity between the predicted characteristics and the real characteristics of land-cover objects in the post-change scene is calculated to identify whether or not any changes occurred. Since the satellite parameters are difficult to acquire in most cases, later work [19] estimated the feature transformation functions directly from remote sensing images via online training based on the *K*-nearest neighbor (KNN) algorithm. The feature transformation method [19] achieve superior performance than [18] in situations where less *a priori* information (e.g., satellite parameters) are available. Note that the computational cost of online training in [19] increases significantly when searching for the *K*-nearest neighbors in massive number of training samples from large remote sensing images [4, 19].

In more recent work, deep neural network (DNN) based methods for the heterogeneous CD task have emerged with the rapid advancement and powerful representation capabilities of DNNs [10]. Refs. [4] and [20] exploit deep convolution networks with a symmetric structure to transform heterogeneous images into a common feature space and extract the bi-temporal features with consistent representations. Refs. [21] employ deep image style transfer networks to convert bi-temporal heterogeneous images into a homogeneous style space, preserving the semantic contents of each image while transforming the texture information of bi-temporal images into a uniform style. In [22-24], generative adversarial networks (GANs) are utilized to translate one image into the domain of the other image, which noticeably reduces the heterogeneity between images of different modalities and enhances the saliency of changes. In addition, the autoencoders in [25] and [26] align the features of bi-temporal heterogeneous images in a latent space via adversarial training to facilitate the learning of the image-to-image translation functions. Although the aforementioned DNN-based methods have made significant progress for the heterogeneous CD task, the operating mechanisms and internal principles of DNNs are often not available to us in most cases, resulting in degraded interpretability[1] in practical CD applications.

Copula theory describes the joint probability distribution of multivariate random variables via the product of a copula density function and their respective marginal density functions, and is an effective tool to model statistical dependence relationships among random variables [27]. Based on our introspection, we believe that the essence of the CD task in our context is to accurately measure the statistical dependence of bi-temporal data. Due to this essence of CD, copula theory is highly appropriate for this task. In [28], copula theory has been applied to capture the dependence between bi-temporal heterogeneous images in unchanged areas, and then the copula-based quantile regression is performed to calculate a simulated statistics at the post-change time. Finally, the changed areas are identified by comparing the simulated statistics with the original statistics in the post-change image. Note that the copula functions used in [28] are manually selected from the existing copula families based on the visual inspection of the dependence of the given bi-temporal images. In practice, the dependence of heterogeneous image pairs varies quite a bit with different scenarios and sensors. Thus, the selection of a suitable copula function from the existing copula families for various scenes is labor-consuming. Besides, most existing copula functions are previously customized for other fields such as finance [29], which also hinders the usability of copula theory in remote sensing tasks.

---
[1] We will give the definition and taxonomy of interpretability in Section II.B.

In recent years, some studies [30-32] have attempted to model copula functions via neural networks. Thanks to the powerful and flexible representation capabilities of neural networks, copula functions learned by neural networks exhibit a comparable or even superior generalization than the traditional copula functions. The intrinsic essence of CD has been demonstrated to have a close relationship with copula theory [28, 33], which naturally provides an opportunity to introduce copula theory into neural networks for CD. Inspired by the insights of [28, 30-33], we propose a novel semi-supervised heterogeneous CD method based on a copula-guided neural network, named NN-Copula-CD, to obtain better interpretability than the existing DNN-based heterogeneous CD methods while enhancing the robustness of traditional copulas applied for remote sensing data. The core idea of our NN-Copula-CD is to construct a neural network to estimate the degrees of dependence between registered regions in bi-temporal images under the guidance of copula theory. First, our NN-Copula-CD utilizes a neural network to learn a copula function from a few unchanged superpixel[2] pairs. The learned copula function extracts the unchanged patterns between bi-temporal images into statistical dependence measurement between two superpixels, where the unchanged superpixel pairs tend to be strongly dependent and the changed superpixel pairs tend to be weakly dependent. Next, we exploit the learned copula function to calculate the degrees of dependence for all superpixel pairs in the bi-temporal images and then the changed and unchanged superpixels can be easily distinguished. Note that compared with most existing CD methods based on DNNs, the neural network in our NN-Copula-CD has better interpretability based on copula theory, i.e., mathematical constraints that a copula function should satisfy are employed as loss functions to guide the training process of neural networks. This encourages the trained neural network to follow the mathematical characteristics of copula theory and to approximate a copula function that reflects the dependence between unchanged regions in bi-temporal heterogeneous images.

---

[2] The notion of superpixels will be defined later in Section III.B.

The main contribution of our work is that, a new knowledge-data-driven CD method based on the copula-guided neural network, i.e., NN-Copula-CD, is proposed to tackle the heterogeneous CD problem. In NN-Copula-CD, we train a neural network under the guidance of copula theory, and then exploit it to model the dependence between the pre-change and post-change superpixels to identify the changed areas. During this process, the generalization capabilities of data-driven neural networks are integrated with the theoretical basis of knowledge-driven copula theory, which improves the interpretability and robustness of the existing DNN-based and traditional-copula-based heterogeneous CD methods, respectively. Experimental results on three heterogeneous CD datasets demonstrate the superiority of the proposed NN-Copula-CD over the commonly used methods in terms of kappa coefficient. Visual results also exhibit the effectiveness of our method.

The rest of this paper is organized as follows. Section II reviews copula theory, the definition and taxonomy of interpretability, and the existing heterogeneous CD methods based on neural networks. Section III presents the formulation and specifics of the proposed NN-Copula-CD method. Section IV provides the in-depth experiments on three heterogeneous CD datasets. Finally, the conclusion and our future directions are given in Section V.

# II. Background and Related Works

In this section, a brief background on copula theory and interpretability along with a more complete discussion on the literature that are directly related to our work are presented.

*A. Copula Theory*

Sklar's theorem [34] indicates that the joint distribution of multivariate random variables can be represented in terms of all the marginal distributions of random variables and a copula function. Let $F(\nu_1, \nu_2, ..., \nu_d)$ denote the joint cumulative distribution function (CDF) of random variables $\nu_1, \nu_2, ..., \nu_d$, and their marginal cumulative distribution functions are represented by $F_1(\nu_1), F_2(\nu_2), ..., F_d(\nu_d)$, respectively. There always exists a *d*-dimensional copula function $C(\cdot)$ [35] where

$$F(\nu_1, \nu_2, ..., \nu_d) = C(F_1(\nu_1), F_2(\nu_2), ..., F_d(\nu_d)). \tag{1}$$

By utilizing an appropriate copula function, Sklar's theorem allows one to handle the dependence between the random variables and the marginal distributions of each random variable [36] separately, which provides a convenient and effective way to deal with complex dependence modeling problems in quantitative finance, signal processing, and remote sensing.

Copula theory has been applied for remote sensing applications in some prior works. Wang *et al*. [37] proposed a copula-based fusion strategy that utilizes clutter correspondence between multi-source synthetic aperture radar (SAR) images to enhance the target detection performance. Mercier *et al*. [28] proposed a CD model that utilizes conditional copulas to generate the simulated statistics from pre-change statistics of observations. These simulated statistics are then compared with the real post-change statistics via a modified Kullback-Leibler divergence to identify the changes. Note that the copula functions used in [28] need to be specified manually. However, since most copula functions are custom-made for finance and other fields, it is difficult to find a generalized copula function to robustly fit remote sensing data of different scenarios. As a result, the CD performance could be seriously degraded in case an inappropriate copula function is chosen.

*B. Interpretability of Machine Learning Methods*

With the swift development of machine learning (ML) methods (especially neural networks) in various fields, the interpretability of data-driven ML methods has gathered increasing attention and discussion from scholars [38-40]. A general definition for interpretability is the capability that offers explanations in understandable terms to make humans interpret the decision-making mechanism of ML methods [39, 40]. Just like humans can explain the same thing from different perspectives, interpretable ML methods can also be divided into various taxonomies with different criteria [39, 41]. Here, we introduce the taxonomy of interpretable ML methods used in [39, 41], where interpretability in ML methods is classified into three categories, i.e., pre-model interpretability, in-model interpretability, and post-model interpretability.

The pre-model interpretability comes from the interpretable operations that preprocess datasets and features before model building and training [41]. The representative case of these interpretable operations is exploratory data analysis, which focus on analyzing correlations and regularity in features and datasets to understand the patterns of inputs for models [39]. The in-model interpretability represents the inherent interpretability of the model and relies on the theoretical rules and constraints used in the process of building and training models [39, 41]. An example of in-model interpretability is physical information neural networks [42] (PINNs), which integrate theoretical rules and mathematical constraints into the training of neural networks to approximate certain physical regulations. These theoretical rules and mathematical constraints are often employed as different loss functions to guide the training of neural networks, where the decision-making process of trained neural networks is potentially supervised by the corresponding physical regulations. Post-model interpretability primarily emerges after the models have been trained [39]. One of the most representative examples for post-model interpretability is the gradient-weighted class activation mapping (Grad-CAM) strategy [43], where the gradients of predictions with respect to the activation map of the last convolution layer are utilized to help interpret the contributions of different parts of the image towards the predictions.

Note that in this paper, we consider the interpretability in the proposed NN-Copula-CD as in-model interpretability because the neural copula model in NN-Copula-CD is trained under the guidance of copula theory, which is similar to the idea of PINN [42].

*C. Neural Networks for Heterogeneous Change Detection*

In the past decade, neural networks have been widely used for various remote sensing tasks [44-46] due to their powerful nonlinear modeling capabilities. In recent CD studies, image/feature domain alignment and dependence modeling have been widely investigated using neural networks for both homogeneous and heterogeneous CD. Regarding the heterogeneous CD task, various neural networks including image-to-image translation networks and deep homogeneous transformation networks have been utilized to address the challenge of heterogeneity. Image-to-image translation networks translate an image from the domain of one temporal image into the domain of the other temporal image while maintaining the semantic content [47]. Deep homogeneous transformation networks simultaneously transform bi-temporal images into a deep feature space where the feature representations of two images are approximately homogeneous [20].

For the image-to-image translation networks [22-25], Niu *et al*. [24] employed the conditional generative adversarial network (cGAN) to translate one image into the representation space of the other image, and then directly compared the translated source image and the original target image to obtain the CD results. Refs. [23] and [25] extended Niu's work [24] by improving the cGAN with the idea of cycle-consistent adversarial networks [48] (CycleGANs) that provides superior stability and translation performance. The noise in the difference map between the translated pre-change image and the original post-change image is also considered in [23], where a random forest classifier replaces direct comparison to improve the robustness of CD results. In addition, Refs. [21] leveraged the different layers of the visual geometry group (VGG) network to extract semantic features and style features, and then transferred the style features of bi-temporal images into the homogeneous space while keeping semantic features the same across bi-temporal images. Then, the changed areas can be easily recognized via a homogeneous CD method.

For the deep homogeneous transformation networks, the coupling network in [20] simultaneously transforms the bi-temporal heterogeneous images into the homogeneous feature space via deep coupling convolution layers, where the extracted deep homogeneous features can be directly used to compare and identify the changes. Furthermore, a Siamese network proposed in [4] focuses on deep-level feature transformation and the efficiency of CD is improved by using fewer network parameters that are required to be optimized.

Note that the existing neural networks [4, 20-25] have delivered outstanding performance for the heterogeneous CD task, but the lack of interpretability therein is still an issue. This deficiency results in limited controllability since the operating mechanism of most existing neural networks for heterogeneous CD is like a black box for researchers. Therefore, in this paper, we attempt to propose a neural network method for heterogeneous CD via the theoretical guidance of copula theory with better interpretability.

# III. Proposed NN-Copula-CD Method

In this section, our proposed NN-Copula-CD method is introduced. We begin by formulating the heterogeneous CD problem and presenting the overall framework of the proposed method. Then, we introduce the notions of superpixel segmentation and training sample selection, the training of the copula-guided interpretable neural network for heterogeneous CD, and the generation of final detection results in our NN-Copula-CD method.

*A. Problem Formulation and Overview of the Approach*

Assume that $I^{T_1}$ and $I^{T_2}$ are bi-temporal co-registered heterogeneous remote sensing images of the same location at the pre-event time $T_1$ and after-event time $T_2$, respectively. The goal of our method is to recognize the changes between $I^{T_1}$ and $I^{T_2}$, and produce pixel-level binary labels to indicate the location of the changes.

We first present a high-level overview of the proposed NN-Copula-CD method as well as definitions in this section. We will provide more details later. In the first step, we segment the bi-temporal images into superpixel pairs and select a few unchanged regions as the training regions. Then, a fully connected neural network is trained under the guidance of copula theory to learn the dependence between the unchanged bi-temporal superpixel pairs involved in the training region. Next, the trained neural copula is exploited to calculate the degrees of dependence for all the superpixel pairs in the bi-temporal images, where the degrees of dependence are high(low) in the unchanged(changed) regions. Finally, the fuzzy *c*-means algorithm [49] is utilized to classify all the superpixel pairs into the changed and unchanged categories according to the calculated degrees of dependence.

The overall framework of the proposed method is shown in Fig. 1. Our method is semi-supervised and the training process can be divided into three steps (see the blue stream in Fig. 1), i.e., superpixel segmentation, training sample selection, and neural copula training. The test process also consists of three steps (see the red stream in Fig. 1), i.e., superpixel segmentation, neural copula inference, and clustering and classification using the fuzzy *c*-means algorithm.

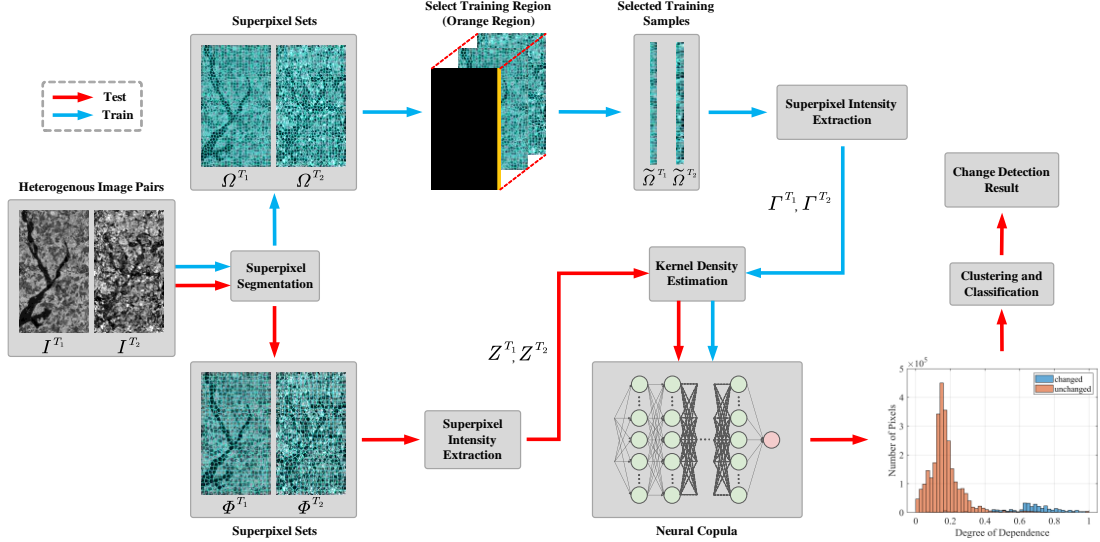

Fig. 1. Overall framework of the proposed NN-Copula-CD.

*B. Superpixel Segmentation and Training Sample Selection*

In most remote sensing applications, the acquired images are quite large. In addition, heterogeneity caused by different sensors also seriously challenges the pixel-level dependence analysis. Therefore, it may not be efficient to directly build a pixel-to-pixel dependence relationship for high-resolution bi-temporal heterogeneous remote sensing images [12, 13]. Compared with individual pixels, superpixels are the segmentation results of perceptual clustering algorithms that group adjacent pixels with similar visual characteristics [50]. As a result, superpixels provide a more comprehensive representation of a group of pixels, and are commonly used as basic units for heterogeneous CD analysis [11-13, 51]. In the proposed NN-Copula-CD method, a collaborative simple linear iterative clustering algorithm (Co-SLIC) in [13] is applied to segment heterogeneous image pairs into superpixel pairs, which customizes the local clustering rules for different modalities and achieves the co-segmentation for bi-temporal heterogeneous images. The superpixel sets of $I^{T_1}$ and $I^{T_2}$ after Co-SLIC segmentation are denoted by $\Omega^{T_1}$ and $\Omega^{T_2}$, respectively, and we use $N_1$ to denote the number of segmented superpixels in both $I^{T_1}$ and $I^{T_2}$. $\Omega_i^{T_1}$ and $\Omega_i^{T_2}$ are the *i*-th superpixels in $\Omega^{T_1}$ and $\Omega^{T_2}$, respectively.

Once the superpixel segmentation is completed, the next step is to train the neural copula model. Due to the inconsistent distributions of heterogeneous data across different scenes, training a neural copula model offline from other scenes using historical data often leads to inferior CD performance. In light of this, we choose a few unchanged superpixel pairs directly from the original images as training samples to train the neural copula model online, like the existing heterogeneous CD methods in [4, 19]. However, the online training regions are manually picked by professionals in [4, 19] and the training sample selection process therein is time-consuming and labor-intensive. In most emergencies, satellites will ensure that the coverage of the captured images is wide enough to fully cover the area of change that is of our interest (such as the epicenter) [52], which means that the captured images usually contain unchanged areas at the edge of the large satellite images. Thus, we follow a manner similar in [33] and choose to select a region nearby the edge of the input bi-temporal images as the selected training region (see the orange region as shown in Fig. 1). This strategy not only ensures the consistency of the distribution between the training samples and the original data, but also alleviates time and labor costs to select training samples. Superpixels used for training will be picked from $\Omega^{T_1}$ and $\Omega^{T_2}$ based on whether the intersection of pixels between the current superpixel and the selected training region is larger than half of the pixels within the current superpixel. Suppose that $\Lambda$ is the selected training region, the process of superpixel selection for training is expressed as:

$$\begin{aligned}\widetilde{\Omega}^{T_1} &= \{\Omega_j^{T_1} | \frac{Num(\Omega_j^{T_1} \cap \Lambda)}{Num(\Omega_j^{T_1})} > 0.5, \Omega_j^{T_1} \in \Omega^{T_1}, 1 \leq j \leq N_1\},\\ \widetilde{\Omega}^{T_2} &= \{\Omega_j^{T_2} | \frac{Num(\Omega_j^{T_2} \cap \Lambda)}{Num(\Omega_j^{T_2})} > 0.5, ,\Omega_j^{T_2} \in \Omega^{T_2}, 1 \leq j \leq N_1\},\end{aligned} \quad (2)$$

where $Num(\cdot)$ is the function that counts the number of pixels. $\widetilde{\Omega}^{T_1}$ and $\widetilde{\Omega}^{T_2}$ are the selected training superpixel sets both with $N_2$ superpixels from $I^{T_1}$ and $I^{T_2}$, respectively. We use the mean value of all the pixels within a training superpixel, i.e., superpixel intensity, to concisely represent the current superpixel, i.e.,

$$\begin{aligned}\Gamma^{T_1} &= \left\{\gamma_k^{T_1} = \frac{Floor\left(Mean\left(\widetilde{\Omega}_k^{T_1}\right)\right)}{255}, 1 \leq k \leq N_2\right\},\\ \Gamma^{T_2} &= \left\{\gamma_k^{T_2} = \frac{Floor\left(Mean\left(\widetilde{\Omega}_k^{T_2}\right)\right)}{255}, 1 \leq k \leq N_2\right\},\end{aligned} \quad (3)$$

where $Mean(\cdot)$ is the function that calculates the mean value of all the pixels in the corresponding superpixel, and $Floor(\cdot)$ is used to truncate the decimal parts of numbers. Since the range of pixel values in most remote sensing images is 0 to 255, we normalize the superpixel intensity via a division by 255. $\Gamma^{T_1}$ and $\Gamma^{T_2}$ are the sets containing the normalized superpixel intensity values of training superpixels in $\widetilde{\Omega}^{T_1}$ and $\widetilde{\Omega}^{T_2}$, respectively. It is worth mentioning that there might be more features (e.g., variance and median value) available to represent the characteristics of superpixels. In this paper, we only utilize the mean values to show the effectiveness of our copula-guided CD mechanism since our focus is not to exhaustively exploit the features of superpixels. We refer the readers to [12, 13] for more superpixel features used in the CD task.

*C. Neural Copula Model*

Once the training superpixel sets $\Gamma^{T_1}$ and $\Gamma^{T_2}$ are prepared, the next step is to model the dependence relationship between the unchanged areas of $\Gamma^{T_1}$ and $\Gamma^{T_2}$ via copula functions. Traditional copula functions are designed based on *a priori* knowledge of data distributions, focusing on the specific dependence relationships of interest [53]. For example, Clayton copula focuses on negative tail dependence while Frank copula is concerned about symmetric tail dependence. However, *a priori* knowledge of the distribution of remote sensing data is difficult to know precisely due to the various imaging conditions and scenarios. Therefore, we are often uncertain about which traditional copula function is the most appropriate in advance. To address this issue, we adopt a new neural copula model in [32] to learn the dependence between unchanged regions from remote sensing data.

The first step of copula theory is to impose the probability integral transform for each marginal distribution, i.e., to estimate the marginal CDF values for every element in $\Gamma^{T_1}$ and $\Gamma^{T_2}$. In our implementation, we utilize the kernel density estimation approach (KDE) [54] to estimate the marginal CDF values. Suppose that $KDE(Y;X)$ denotes the function that estimates the marginal CDF values at every element in the set $Y$ according to the distribution fitted by elements in the set $X$. The estimation problem for marginal CDF values in $\Gamma^{T_1}$ and $\Gamma^{T_2}$ can be formulated as:

$$\begin{aligned}\tilde{\boldsymbol{u}} &= KDE(\Gamma^{T_1};\Gamma^{T_1}),\\ \tilde{\boldsymbol{v}} &= KDE(\Gamma^{T_2};\Gamma^{T_2}),\end{aligned} \quad (4)$$

where $\tilde{\boldsymbol{u}}$ and $\tilde{\boldsymbol{v}}$ are the marginal CDF values estimated from $\Gamma^{T_1}$ and $\Gamma^{T_2}$, respectively. Since the normalized superpixel intensities in $\Gamma^{T_1}$ and $\Gamma^{T_2}$ range from 0 to 1 with 256 evenly spaced elements, we exploit KDE to estimate the marginal CDF values at these evenly spaced elements and then construct a CDF value table. Next, we obtain the marginal CDF values for every element in $\varGamma^{T_1}$ and $\varGamma^{T_2}$ by table lookup of the established CDF value table. This lookup-table operation can significantly alleviate duplicate computation in the marginal CDF estimation process.

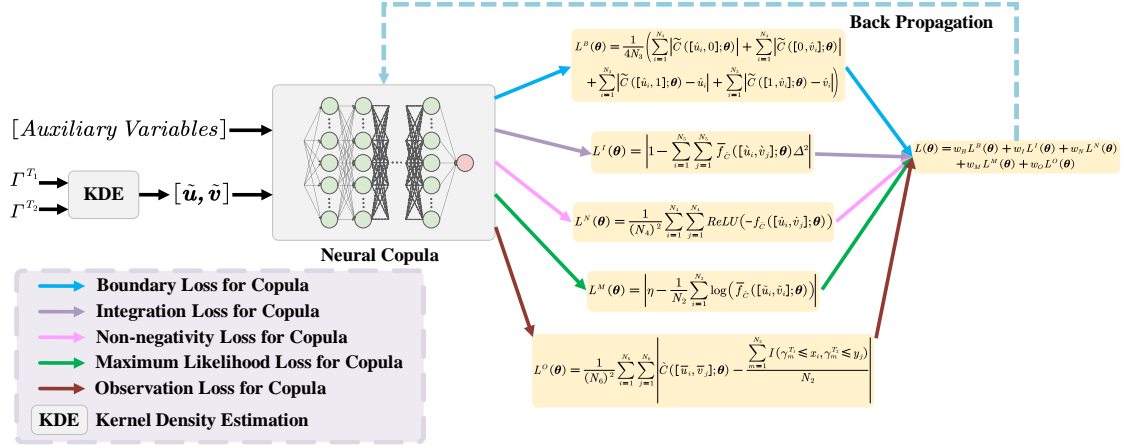

Fig. 2. Training diagram of the neural copula in our NN-Copula-CD. The auxiliary variables contain the vectors $[\hat{\boldsymbol{u}},\hat{\boldsymbol{v}}]$, $[\dot{\boldsymbol{u}},\dot{\boldsymbol{v}}]$, $[\ddot{\boldsymbol{u}},\ddot{\boldsymbol{v}}]$, and $[\overline{\boldsymbol{u}},\overline{\boldsymbol{v}}]$ that facilitate the neural copula model to satisfy the boundary constraint, integration constraint, non-negativity constraint, and observation constraint, respectively.

Next, we introduce the network structure and the training process of the neural copula. As shown in Fig. 2, the neural copula model is a fully connected neural network that includes three parts, i.e., input layer, hidden layer, and output layer. In our implementation, we construct the fully connected neural network with five hidden layers and twenty neurons per layer as the neural copula model. Suppose that the network is written as $\widetilde{C}([\boldsymbol{u},\boldsymbol{v}];\boldsymbol{\theta})$, where $\boldsymbol{\theta}$ denotes the trainable parameters. The input CDF values are $[\boldsymbol{u},\boldsymbol{v}]$, where $\boldsymbol{u}$ and $\boldsymbol{v}$ are the variables representing the CDF values of the marginal distributions, respectively. The outputs of the network provide the values of the copula function and the gradient $f_{\widetilde{C}}([\boldsymbol{u},\boldsymbol{v}];\boldsymbol{\theta})$ denotes the copula probability density function (PDF). Formally,

$$f_{\widetilde{C}}([\boldsymbol{u},\boldsymbol{v}];\boldsymbol{\theta}) = \frac{\widetilde{C}([\boldsymbol{u},\boldsymbol{v}];\boldsymbol{\theta})}{\partial \boldsymbol{u}\, \partial \boldsymbol{v}}, \qquad (5)$$

Afterward, we tap the potential of the neural network for the CD task by training it under the guidance of copula theory. During the training process, we push the fully connected neural network to converge into a copula function via five mathematical loss functions.

*1) Boundary Constraint of Neural Copula*

The distribution of $\widetilde{C}([\boldsymbol{u},\boldsymbol{v}];\boldsymbol{\theta})$ should obey the boundary constraint of copula's definition, i.e.,

$$\begin{cases} \widetilde{C}([\hat{\boldsymbol{u}},\boldsymbol{0}];\boldsymbol{\theta}) = \boldsymbol{0} \\ \widetilde{C}([\boldsymbol{0},\hat{\boldsymbol{v}}];\boldsymbol{\theta}) = \boldsymbol{0} \\ \widetilde{C}([\hat{\boldsymbol{u}},\boldsymbol{1}];\boldsymbol{\theta}) = \hat{\boldsymbol{u}} \\ \widetilde{C}([\boldsymbol{1},\hat{\boldsymbol{v}}];\boldsymbol{\theta}) = \hat{\boldsymbol{v}} \end{cases} \quad \hat{\boldsymbol{u}} = [\hat{u}_1,\hat{u}_2,...,\hat{u}_{N_3}] \in [0,1] \atop \hat{\boldsymbol{v}} = [\hat{v}_1,\hat{v}_2,...,\hat{v}_{N_3}] \in [0,1] , \tag{6}$$

where $\hat{\boldsymbol{u}} = [\hat{u}_1,\hat{u}_2,...,\hat{u}_{N_3}]$ and $\hat{\boldsymbol{v}} = [\hat{v}_1,\hat{v}_2,..,\hat{v}_{N_3}]$ are two vectors both with $N_3$ evenly spaced elements between 0 and 1. $[\hat{\boldsymbol{u}},\boldsymbol{0}]$, $[\hat{\boldsymbol{u}},\boldsymbol{1}]$, $[\boldsymbol{0},\hat{\boldsymbol{v}}]$, and $[\boldsymbol{1},\hat{\boldsymbol{v}}]$ are the boundary point sets of the copula function. Thus, the boundary loss for the neural copula is defined as:

$$L^B(\boldsymbol{\theta}) = \frac{1}{4N_3} \Bigg( \sum_{i=1}^{N_3} \big| \widetilde{C}([\hat{u}_i,0];\boldsymbol{\theta}) \big| + \sum_{i=1}^{N_3} \big| \widetilde{C}([0,\hat{v}_i];\boldsymbol{\theta}) \big| \\ + \sum_{i=1}^{N_3} \big| \widetilde{C}([\hat{u}_i,1];\boldsymbol{\theta}) - \hat{u}_i \big| + \sum_{i=1}^{N_3} \big| \widetilde{C}([1,\hat{v}_i];\boldsymbol{\theta}) - \hat{v}_i \big| \Bigg). \tag{7}$$

*2) Non-Negativity Constraint of Neural Copula*

The PDF values of all the points in the copula function domain should be non-negative. In the training process, we need to punish the case that the copula PDF values are negative. When copula PDF is continuous, the non-negativity constraint can be formulated as:

$$\int_0^1 \int_0^1 ReLU\big(-f_{\tilde{C}}([\boldsymbol{u},\boldsymbol{v}];\boldsymbol{\theta})\big) d\boldsymbol{u} d\boldsymbol{v} = 0. \tag{8}$$

Similarly, suppose that $\dot{\boldsymbol{u}} = [\dot{u}_1,\dot{u}_2,...,\dot{u}_{N_4}]$ and $\dot{\boldsymbol{v}} = [\dot{v}_1,\dot{v}_2,...,\dot{v}_{N_4}]$ are two vectors both with $N_4$ evenly spaced elements between 0 and 1. The non-negativity loss can be rewritten as:

$$L^N(\boldsymbol{\theta}) = \frac{1}{(N_4)^2} \sum_{i=1}^{N_4} \sum_{j=1}^{N_4} ReLU\big(-f_{\tilde{C}}([\dot{u}_i,\dot{v}_j];\boldsymbol{\theta})\big). \tag{9}$$

*3) Integration Constraint of Neural Copula*

The integration of PDF in the copula function domain should be 1. When the copula PDF is continuous, the integration constraint can be formulated as:

$$\int_0^1 \int_0^1 f_{\tilde{C}}([\boldsymbol{u},\boldsymbol{v}];\boldsymbol{\theta}) d\boldsymbol{u} d\boldsymbol{v} = 1. \tag{10}$$

We adopt the following steps used in [32] to approximately implement the constraint of Eq. (10). Firstly, to prevent the PDF of neural copula from being negative, we need to conduct a non-negative processing operation for the gradient $f_{\widetilde{C}}([\boldsymbol{u},\boldsymbol{v}],\boldsymbol{\theta})$. Formally,

$$\overline{f}_{\widetilde{C}}([\boldsymbol{u},\boldsymbol{v}];\boldsymbol{\theta}) = ReLU\big(f_{\widetilde{C}}([\boldsymbol{u},\boldsymbol{v}];\boldsymbol{\theta})\big) + \rho, \tag{11}$$

where $ReLU(\cdot)$ is the rectified linear unit function [55] and $\rho$ is a minuscule positive value (set to $10^{-9}$ in our experiments). $\rho$ is set to ensure that the input values of the *log* function in Eq. (13) are positive. Secondly, to implement the integral of PDFs corresponding to the discrete pixel values in remote sensing images, Eq. (10) is rewritten in an approximate form. Suppose that $\grave{\boldsymbol{u}} = [\grave{u}_1, \grave{u}_2, ..., \grave{u}_{N_5}]$ and $\grave{\boldsymbol{v}} = [\grave{v}_1, \grave{v}_2, ..., \grave{v}_{N_5}]$ are two vectors both with $N_5$ evenly spaced elements between 0 and 1, and $\Delta = 1/(N_5 - 1)$ is the step size. The integration loss from Eq. (10) in the discrete case can be approximated as:

$$L^I(\boldsymbol{\theta}) = \left| 1 - \sum_{i=1}^{N_5} \sum_{j=1}^{N_5} \overline{f}_{\widetilde{C}}([\grave{u}_i, \grave{v}_j];\boldsymbol{\theta}) \Delta^2 \right|. \tag{12}$$

*4) Maximum Likelihood Estimation for Neural Copula*

To model the dependence relationships in heterogeneous image pairs, the output of the neural copula model needs to approach the distribution of $\tilde{\boldsymbol{u}}$ and $\tilde{\boldsymbol{v}}$ acquired from $\varGamma^{T_1}$ and $\varGamma^{T_2}$ (see Eq. (4)) as closely as possible. This is defined using maximum likelihood estimation (MLE). The maximum likelihood loss is defined as follows:

$$L^M(\boldsymbol{\theta}) = \left| \eta - \frac{1}{N_2} \sum_{i=1}^{N_2} \log\big(\overline{f}_{\widetilde{C}}([\tilde{u}_i, \tilde{v}_i];\boldsymbol{\theta})\big) \right|. \tag{13}$$

$\eta$ should be a positive finite value in implementation. In our experiments, we have empirically found that the setting of $\eta = 10$ is satisfactory.

*5) Observation Constraint of Neural Copula*

To ensure consistency between the CDFs of the neural copula and the empirical CDF of the original bi-temporal data (i.e., $\varGamma^{T_1}$ and $\varGamma^{T_2}$), we need to sample some observation points and narrow the difference between the neural copula CDF values and the empirical CDF values at these points. Suppose that $\boldsymbol{x} = [x_1, x_2, ..., x_{N_6}]$ and $\boldsymbol{y} = [y_1, y_2, ..., y_{N_6}]$ are two vectors both with $N_6$ evenly spaced elements between 0 and 1. The observation points can be sampled as $(x_i, y_j)$ where $i$ and $j$ both ranges from 1 to $N_6$. The marginal CDF values of $\boldsymbol{x}$ and $\boldsymbol{y}$ generated from the distribution of $\varGamma^{T_1}$ and $\varGamma^{T_2}$ are $\overline{\boldsymbol{u}}$ and $\overline{\boldsymbol{v}}$, respectively,

$$\begin{aligned} \overline{\boldsymbol{u}} &= KDE(\boldsymbol{x}; \varGamma^{T_1}), \\ \overline{\boldsymbol{v}} &= KDE(\boldsymbol{y}; \varGamma^{T_2}), \end{aligned} \tag{14}$$

the observation loss can be written as:

$$L^O(\boldsymbol{\theta}) = \frac{1}{(N_6)^2} \sum_{i=1}^{N_6} \sum_{j=1}^{N_6} \left| \tilde{C}([\bar{u}_i, \bar{v}_j]; \boldsymbol{\theta}) - \frac{\sum_{m=1}^{N_2} I\left(\gamma_m^{T_1} \leqslant x_i, \gamma_m^{T_2} \leqslant y_j\right)}{N_2} \right|., \quad (15)$$

where $I(\cdot)$ represents the indicator function,

$$I(z) = \begin{cases} 1 & \text{if } z \text{ is true} \\ 0 & \text{if } z \text{ is not true} \end{cases}. \quad (16)$$

Finally, we combine the aforementioned five constraint loss terms to form the overall loss function,

$$L(\boldsymbol{\theta}) = w_B L^B(\boldsymbol{\theta}) + w_I L^I(\boldsymbol{\theta}) + w_N L^N(\boldsymbol{\theta}) + w_M L^M(\boldsymbol{\theta}) + w_O L^O(\boldsymbol{\theta}), \quad (17)$$

where we set the weights $[w_B, w_I, w_N, w_M, w_O]$ as $[2, 0.3, 1, 0.1, 5]$ in our experiments because they are empirically found to be robust in most scenarios.

*D. Use of the Trained Model to Obtain CD Results*

Once the training of the neural copula model has been completed, we utilize the trained model to infer change areas between $I^{T_1}$ and $I^{T_2}$. We refer to the use of the trained model to generate CD results as the testing stage. The first step of the testing stage is superpixel segmentation. Note that the superpixel segmentation results from the training stage can be directly used here to avoid redundant computation. However, during the testing stage, we once again apply Co-SLIC to segment the original images $I^{T_1}$ and $I^{T_2}$ into distinct superpixel sets, which may result in a different number of superpixels (denoted as $N_7$). The purpose for this repetition of superpixel segmentation is to ensure the superpixel scales in training and testing stages are different so that the trained neural copula model does not show obvious over-fitting performance on a single superpixel scale. The segmented superpixel sets of the original images $I^{T_1}$ and $I^{T_2}$ generated in the testing stage are denoted by $\Phi^{T_1}$ and $\Phi^{T_2}$, respectively. Then, similar to Eq. (3), the normalized superpixel intensity values for each superpixel in $\Phi^{T_1}$ and $\Phi^{T_2}$ are calculated, and are denoted by $Z^{T_1}$ and $Z^{T_2}$, respectively.

The next step is to transform $Z^{T_1}$ and $Z^{T_2}$ into probability space, i.e., to estimate the marginal CDF values for each element in $Z^{T_1}$ and $Z^{T_2}$. We calculate the marginal CDF values of each element in $Z^{T_1}$ and $Z^{T_2}$ according to the marginal distributions fitted by the training sets $\Gamma^{T_1}$ and $\Gamma^{T_2}$, respectively,

$$\ddot{u} = KDE(Z^{T_1}; \Gamma^{T_1}),$$
$$\ddot{v} = KDE(Z^{T_2}; \Gamma^{T_2}), \qquad (18)$$

where $\ddot{u}$ and $\ddot{v}$ are the marginal CDF values of $Z^{T_1}$ and $Z^{T_2}$, respectively. Then, we input $\ddot{u}$ and $\ddot{v}$ into the trained copula model and calculate the PDF values of the neural copula model via differentiation (similar to Eq.(5)):

$$\overline{f}_{\widetilde{C}}([\ddot{u},\ddot{v}];\boldsymbol{\theta}) = ReLU\left(\frac{\widetilde{C}([\ddot{u},\ddot{v}];\boldsymbol{\theta})}{\partial \ddot{u} \partial \ddot{v}}\right) + \rho, \qquad (19)$$

where $\overline{f}_{\widetilde{C}}([\ddot{u},\ddot{v}];\boldsymbol{\theta})$ represents the degrees of dependence between superpixel pairs in bi-temporal images. An element in $\overline{f}_{\widetilde{C}}([\ddot{u},\ddot{v}];\boldsymbol{\theta})$ with a smaller value means that the dependence relationship of the corresponding superpixel pair is different from the dependence relationships of those unchanged superpixel pairs used for training. Then, we apply a negative logarithmic operation on the degrees of dependence $\overline{f}_{\widetilde{C}}([\ddot{u},\ddot{v}];\boldsymbol{\theta})$ to further enhance the discrimination between changed and unchanged superpixels. Afterward, we use the fuzzy *c*-means clustering algorithm [49] to divide the values of $-\log_{10}\left(\overline{f}_{\widetilde{C}}([\ddot{u},\ddot{v}];\boldsymbol{\theta})\right)$ into two categories, i.e., classifying the superpixels into the changed and unchanged regions. Finally, we can obtain the CD results by assigning the categories of the superpixels to the corresponding pixels in heterogeneous bi-temporal images.

We summarize the CD process of the proposed NN-Copula-CD in Algorithm 1.

---

**Algorithm 1.** NN-Copula-CD

---

***Input:***
    Bi-temporal heterogeneous images $I^{T_1}$ and $I^{T_2}$.

***Step 1: Superpixel Segmentation and Training Sample Selection***
    1) Segment heterogeneous image pairs $I^{T_1}$ and $I^{T_2}$ into superpixel pairs $\Omega^{T_1}$ and $\Omega^{T_2}$, respectively.
    2) Select a set of unchanged superpixel pairs $\widetilde{\Omega}^{T_1}$ and $\widetilde{\Omega}^{T_2}$ from $\Omega^{T_1}$ and $\Omega^{T_2}$, respectively, see Eq. (2).
    3) Calculate the normalized superpixel intensity $\Gamma^{T_1}$ and $\Gamma^{T_2}$ of $\widetilde{\Omega}^{T_1}$ and $\widetilde{\Omega}^{T_2}$, respectively, see Eq. (3).

***Step 2: Neural Copula Model Training***
    4) Estimate the marginal CDF values $\tilde{u}$ and $\tilde{v}$ with respect to $\Gamma^{T_1}$ and $\Gamma^{T_2}$, respectively, see Eq. (4).
    5) Train the neural copula model via loss functions (see Eq. (7), (9), (12), (13), and (15)) designed according to copula theory.

***Step 3: Obtain CD Results***

6) Similar to 1)-3), segment heterogeneous image pairs $I^{T_1}$ and $I^{T_2}$ into superpixel pairs $\Phi^{T_1}$ and $\Phi^{T_2}$, respectively. Then, calculate the normalized superpixel intensity $Z^{T_1}$ and $Z^{T_2}$ of $\Phi^{T_1}$ and $\Phi^{T_2}$, respectively.

7) Estimate the marginal CDF values $\ddot{u}$ and $\ddot{v}$ with respect to $Z^{T_1}$ and $Z^{T_2}$, respectively.

8) Input $\ddot{u}$ and $\ddot{v}$ to the trained model and then obtain degrees of dependence $\overline{f}_{\widetilde{C}}([\ddot{u},\ddot{v}];\boldsymbol{\theta})$.

9) Apply fuzzy $c$-means clustering algorithm to classify $-\log_{10}\left(\overline{f}_{\widetilde{C}}([\ddot{u},\ddot{v}];\boldsymbol{\theta})\right)$ into changed and unchanged categories.

*Output:*
    Binary map that indicates CD results.

# IV. Experimental Results

In this section, we apply the proposed NN-Copula-CD algorithm on three heterogeneous datasets and compare its performance with that of several state-of-the-art (SOTA) heterogeneous CD methods. In addition, we conduct in-depth discussions to investigate the factors that influence the performance of our NN-Copula-CD algorithm.

In the proposed NN-Copula-CD method, it is necessary to preprocess input images over multiple channels into a single channel form. The channel-dimension reduction preprocessing can significantly facilitate the efficiency of learning the major dependence relationships between heterogeneous image pairs with acceptable information loss. Consequently, input RGB images will be transformed into a grayscale form. Multispectral images will undergo dimensionality reduction using the PCA algorithm [56].

*A. Dataset*

As shown in Fig. 3, the first dataset is of flood areas in Gloucester, England. It includes a pre-change SAR image acquired by the European remote sensing (ERS) satellite in 1999 and a post-change Normalized Difference Vegetation Index (NDVI) image acquired in 2000 [19]. In Fig. 4, the second dataset is still of flood areas in Gloucester, where the pre-change multispectral image is acquired by the Satellites Pour l'Observation de la Terre (SPOT) satellite in 1999 and the post-change NDVI image is captured in 2000 [19]. We follow the image size of this flood scene as in [4]. The sizes of the heterogeneous images in Fig. 3 and Fig. 4 are both 2359×1318.

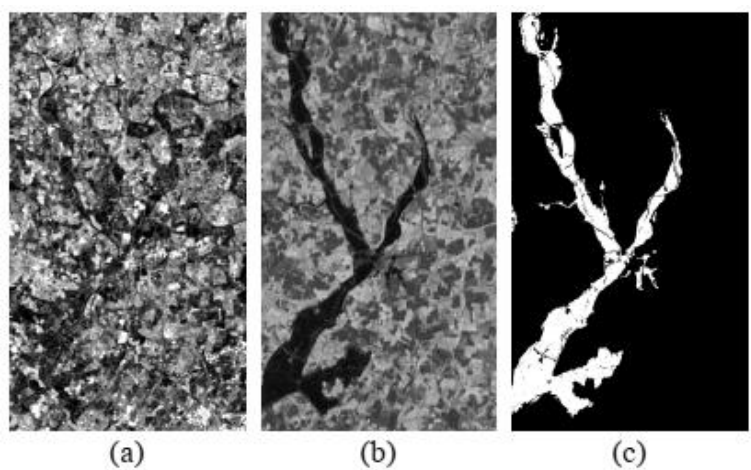

Fig. 3. First dataset. (a) The pre-change SAR image was acquired in 1999. (b) The post-change NDVI image was acquired in 2000. (c) Ground truth.

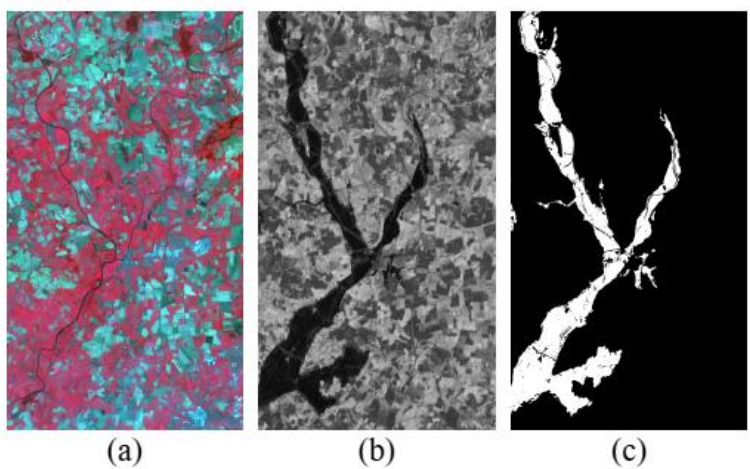

Fig. 4. Second dataset. (a) The pre-change SPOT image was acquired in 1999. (b) The post-change NDVI image was acquired in 2000. (c) Ground truth.

The third dataset in Fig. 5 is provided by [57] and it reflects the urban construction in Toulouse, from 2009 to 2013. This dataset is composed of a pre-change SAR image acquired by the TerraSAR-X satellite in 2009 and a post-change optical image acquired by the Pleiades satellite in 2013. The optical image is originally provided in grayscale form with the size 4404×2600 and the size of the SAR image is 4404×2604. Next, we crop the SAR image into the size 4404×2600, which is the same as the optical image.

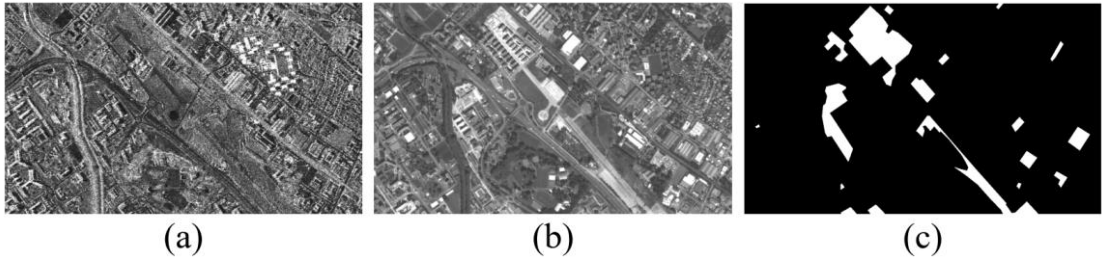

Fig. 5. Third dataset. (a) The pre-change SAR image acquired in 2009. (b) The post-change optical image was acquired in 2013. (c) Ground truth.

*B. Experimental Setting*

*1) Competitors for Heterogeneous CD*

In our experiments, we compare the proposed NN-Copula-CD method with several representative SOTA methods[3] for heterogeneous CD.

HPT [19] is a semi-supervised method based on homogeneous pixel transformation. It transforms one image into the space of the other image in both forward and backward forms, which aligns the pixel-level features of the two images into a homogeneous space. Then, the CD results are generated via the calculation of feature differences and the fuzzy *c*-means clustering algorithm.

SCCN [20] is a well-known unsupervised method based on the deep convolutional coupling network. First, SCCN exploits two denoising autoencoders (DAEs) to initialize each side of the network. Then, the two sides of the network are collaboratively trained to transform the two images in the common deep feature space. Finally, the CD results are obtained by classifying the difference map of the two sides of the network using a thresholding algorithm.

IRG-McS [13] is an unsupervised method based on graph representation and Markov co-segmentation. The IRG-McS method first builds KNN graph representations for bitemporal heterogeneous images, and then utilizes graph mapping to calculate the difference maps of bitemporal graph representations. Finally, the CD results are generated from the difference maps via a Markov co-segmentation model.

*2) Performance Indicators*

We adopt typical CD performance indicators in our experiments, i.e., F1-score (F1) and kappa coefficient (KC) to evaluate the performance of the proposed NN-Copula-CD method and competitors. The definitions of F1 and KC are as follows:

$$\begin{aligned} \text{P} &= \text{TP} / (\text{TP} + \text{FP}), \text{R} = \text{TP} / (\text{TP} + \text{FN}), \\ \text{F1} &= 2\text{PR} / (\text{P} + \text{R}), \\ \text{KC} &= (\text{PCC} - \text{PRE}) / (1 - \text{PRE}), \\ \text{PRE} &= \frac{(\text{TP} + \text{FN})(\text{TP} + \text{FP}) + (\text{TN} + \text{FP})(\text{TN} + \text{FN})}{(\text{TP} + \text{TN} + \text{FP} + \text{FN})^2}, \end{aligned} \qquad (20)$$

where TP, TN, FP, and FN denote the numbers of true positives, true negatives, false positives, and false negative predictions, respectively.

---

[3] We re-implemented HPT and SCCN based on the description provided in their papers, and re-implemented IRG-McS using the public code available at https://github.com/yulisun/IRG-McS.

*3) Implementation Details*

Our NN-Copula-CD method is implemented via the Keras framework [58]. We train our NN-Copula-CD model on NVIDIA RTX TITAN GPU. In fact, the computational burden needed to train our NN-Copula model is very small since it only contains five hidden layers. In the training process, the total training epochs are set to 25000. When the training process is over, we chose the model with the least loss $L(\boldsymbol{\theta})$ as the testing model to produce the CD results. $N_1$ and $N_2$ are set to 6000 and 3000, respectively, for the first and second datasets, while 8000 and 3000 respectively for the third dataset.

*C. Performance Comparison*

The quantitative performance of different methods on the three datasets is reported in Table I. Note that we exploit the same selected unchanged samples to train the semi-supervised methods including HPT and our method. The quantitative comparison in Table I shows that the proposed NN-Copula-CD algorithm achieves a higher score of KC on all three datasets. The semi-supervised competitor HPT obtains the second-best performance in the first flood scene, yet it exhibits a degraded performance in the city scene. The reason for this is that the homogeneous pixel transformation used in the city scene suffers from a much more complex heterogeneity than that in the flood scene, especially for SAR images that contain serious speckle noise. Besides, the categories of land-cover objects in the city scene are more diverse and the same buildings often have significantly different characteristics in optical and SAR images. The unsupervised DNN-based SCCN also achieves good performance in the flood scenes, yet it struggles to achieve robust detection in the city scene. Compared with the existing SOTA competitors, the proposed NN-Copula-CD method significantly reduces the missed predictions and the changed areas identified by the proposed NN-Copula-CD are more complete.

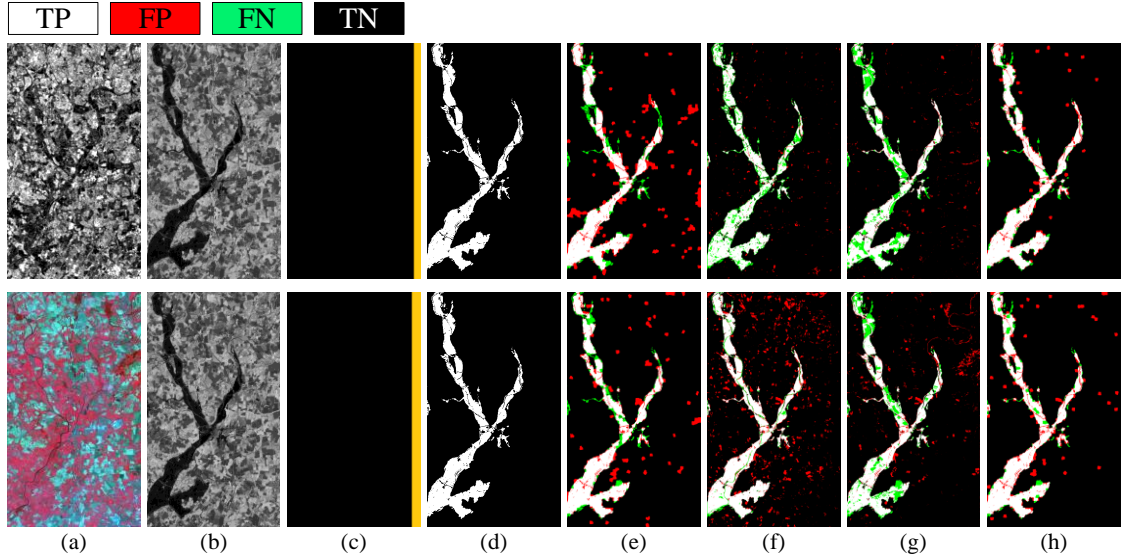

Fig. 6. Visual results of HPT, SCCN, IRG-McS, and NN-Copula-CD on the first and second datasets. (a) The image before change. (b) The image after change. (c) The orange marked regions to train the semi-supervised methods. (d) Change Label. (e) IRG-McS. (f) HPT. (g) SCCN. (h) Our NN-Copula-CD.

TABLE I

COMPARISON RESULTS OF DIFFERENT METHODS ON THREE DATASETS.

| Method | First Dataset | Second Dataset | Third Dataset |
|---|---|---|---|
| | F1 / KC (%) | F1 / KC (%) | F1 / KC (%) |
| IRG-McS | 75.03 / 71.24 | 78.81 / 75.73 | 33.84 / 28.26 |
| SCCN | 77.65 / 75.02 | 77.13 / 74.05 | 21.97 / 14.52 |
| HPT | 82.48 / 80.34 | 78.20 / 74.74 | 27.29 / 22.60 |
| NN-Copula-CD | 86.50 / 84.60 | 82.62 / 80.07 | 40.49 / 33.22 |

TABLE II

COMPARISON RESULTS BETWEEN NEURAL COPULA IN OUR NN-COPULA-CD AND TRADITIONAL COPULAS ON THREE DATASETS.

| Method | First Dataset | Second Dataset | Third Dataset |
|---|---|---|---|
| | F1 / KC (%) | F1 / KC (%) | F1 / KC (%) |
| Gaussian Copula | 71.79 / 68.92 | 81.05 / 78.78 | 14.55 / 2.07 |
| Student-$t$ Copula | 73.07 / 70.25 | 82.26 / 79.53 | 37.83 / 30.22 |
| Clayton Copula | 52.38 / 49.03 | 54.81 / 51.39 | 16.41 / 2.40 |
| Frank Copula | 63.16 / 57.60 | 70.90 / 66.25 | 17.02 / 4.12 |
| NN-Copula-CD | 86.50 / 84.60 | 82.62 / 80.07 | 40.49 / 33.22 |

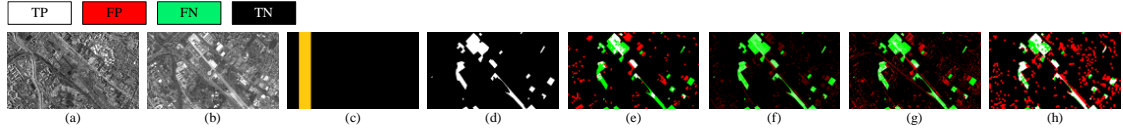
(a) (b) (c) (d) (e) (f) (g) (h)

Fig. 7. Visualization results of HPT, SCCN, IRG-McS, and NN-Copula-CD on the third dataset. (a) The image before change. (b) The image after change. (c) The orange marked training regions for the semi-supervised methods. (d) Change Label. (e) IRG-McS. (f) HPT. (g) SCCN. (h) Our NN-Copula-CD.

## D. Discussion

**Traditional copula functions *vs* neural copula model.** To demonstrate the superiority of the neural copula compared with the traditional copula functions, while performing the heterogeneous CD task we replace the neural copula model with commonly used traditional copula functions, i.e., Student-*t* copula, Gaussian copula, Clayton copula, and Frank copula. Note that the references [28] manually choose the best traditional copula from various traditional copulas via the visual observation of dependencies generated from a graphical method [59]. However, the visual dependence is usually different for various scenes, where the best traditional copula for various scenes is also different and should be individually selected. In our NN-Copula-CD, the neural network provides more parameters, i. e., degrees of freedom, than the commonly used traditional copulas, when manually searching for the best copula from existing traditional copulas [28] is replaced by learning a copula that is most suited for the given scenes. The quantitative results in Table II show that the neural copula surpasses the commonly used traditional copulas on three datasets. This is because our NN-Copula-CD learns the complex dependence from heterogeneous remote sensing data more accurately.

**Comparison between DNN-based method and our NN-Copula-CD under the same training data for supervised CD.** To fairly compare our NN-Copula-CD and the DNN-based SCCN [20] using the same training data, we use the same unchanged training regions as our NN-Copula-CD to supervise the training of SCCN [20]. Inspired by the experiments conducted in [20] implementing the supervised version of SCCN [20], we implement a semi-supervised version of SCCN [20]. Table III displays the comparison results of our NN-Copula-CD and semi-supervised SCCN. The quantitative results demonstrate that our NN-Copula-CD still achieves superior performance compared with SCCN [20] in terms of the KC performance under the same supervised information on both urban and flood scenes.

TABLE III

COMPARISON RESULTS BETWEEN NN-COPULA-CD AND SEMI-SUPERVISED SCCN.

| Dataset | NN-Copula-CD | Semi-supervised SCCN |
|---|---|---|
| | F1 / KC (%) | F1 / KC (%) |
| First Dataset | 86.50 / 84.60 | 84.44 / 82.28 |
| Third Dataset | 40.49 / 33.22 | 32.64 / 25.74 |

**Effect of Number of Training Samples.** To investigate the impact of different number of training samples on the performance of our NN-Copula-CD algorithm, we conduct the corresponding discussion on the first dataset. We select three training regions containing 5.3%, 9.9%, and 16.6% pixels of the original image size, respectively, to train the neural copula model with other settings remaining fixed. The quantitative results using different number of training samples are shown in Table IV. We can observe that the performance of the NN-Copula-CD method achieves the highest KC of 84.6% using 5.3% training samples. With more training samples, the performance of the NN-Copula-CD method does not show an expected improvement. We speculate that two factors may be contributing to this phenomenon. The first factor is that increasing the number of training samples brings higher modeling challenges, and the fixed neural network model cannot adapt to the growing number of samples. The second factor is that we increase the number of training samples by simply expanding the width of the rigid training regions. Although our process to increase the number of training samples is time-efficient and labor-saving, it may bring more noise and low-quality training samples that degrades the CD performance.

TABLE IV

QUANTITATIVE EVALUATION WITH DIFFERENT NUMBER OF TRAINING SAMPLES.

| Number of Training Samples | NN-Copula-CD |
|---|---|
| | F1 / KC (%) |
| 5.31% | 86.50 / 84.60 |
| 9.94% | 84.68 / 82.58 |
| 16.62% | 82.38 / 80.03 |

**Effect of Trainable Parameters in NN-Copula-CD.** To investigate the effect of the number of trainable parameters in NN-Copula-CD, we have developed three variations of NN-Copula-CD, i.e., the small version with 481 parameters, the medium version with 1021 parameters, and the large version with 1761 parameters. The small, medium, and large models all consist of five hidden layers with 10, 15, and 20 neurons in each layer, respectively. We fix other experimental settings and test the three variations of NN-Copula-CD with 5.3% training samples on the first dataset. As the quantitative comparison shown in Table V indicates, the medium version surpasses the other two variations on both PCC and KC. Specifically, the small version and the large version show slight deviations on KC compared with the medium version.

TABLE V
QUANTITATIVE EVALUATION WITH DIFFERENT NUMBER OF PARAMETERS IN OUR NN-COPULA-CD.

| Model Size | NN-Copula-CD |
|---|---|
| | F1 / KC (%) |
| Small (481 Params) | 85.26 / 83.27 |
| Medium (1021 Params) | 86.73 / 84.86 |
| Large (1761 Params) | 86.50 / 84.60 |

# V. Conclusion

In this paper, we proposed a novel semi-supervised method based on the copula-guided neural network, named NN-Copula-CD, to address the CD problem in heterogeneous remote sensing images. In the proposed NN-Copula-CD, we treated CD as an inference problem in terms of statistical dependence, where the changes can be efficiently identified by comparing the degrees of dependence of remote sensing region pairs. In NN-Copula-CD, the neural network that is trained to measure the statistical dependence of bi-temporal images satisfies the characteristics of a copula function under the framework of copula theory. Therefore, our NN-Copula-CD method is endowed with good interpretability in terms of the CD mechanism using neural networks. In addition, our NN-Copula-CD is more robust to model the dependence of heterogeneous remote sensing data than existing CD methods based on traditional copula functions due to the powerful representation capability of neural networks. The in-depth experiments on three CD datasets with heterogeneous remote sensing images demonstrate the effectiveness of our NN-Copula-CD method in comparison with existing state-of-the-art heterogeneous CD methods, including the DNN-based SCCN. In the future, we will investigate the improvement of our NN-Copula-CD method in complex urban scenarios.

# VI. Reference

[1] W. Li, X. Wang, G. Li, B. Geng, and P. K. Varshney, "Change detection in heterogeneous remote sensing images based on copula-supervised neural networks," *2024 IEEE International Conference on Signal, Information and Data Processing*, submitted, 2024.

[2] A. Asokan and J. Anitha, "Change detection techniques for remote sensing applications: a survey," *Earth Science Informatics*, vol. 12, no. 2, pp. 143-160, 2019.

[3] C. Wu, B. Du, and L. Zhang, "Fully convolutional change detection framework with generative adversarial network for unsupervised, weakly supervised and regional supervised change detection," *IEEE Transactions on Pattern Analysis and Machine Intelligence*, vol. 45, no. 8, pp. 9774-9788, 2023.

[4] X. Jiang, G. Li, X.-P. Zhang, and Y. He, "A semisupervised Siamese network for efficient change detection in heterogeneous remote sensing images," *IEEE Transactions on Geoscience and Remote Sensing*, vol. 60, pp. 1-18, 2022.

[5] Z. Zheng, Y. Zhong, J. Wang, A. Ma, and L. Zhang, "Building damage assessment for rapid disaster response with a deep object-based semantic change detection framework: From natural disasters to man-made disasters," *Remote Sensing of Environment*, vol. 265, p. 112636, 2021.

[6] M. K. Ridd and J. Liu, "A comparison of four algorithms for change detection in an urban environment," *Remote sensing of environment*, vol. 63, no. 2, pp. 95-100, 1998.

[7] C. Benedek, X. Descombes, and J. Zerubia, "Building development monitoring in multitemporal remotely sensed image pairs with stochastic birth-death dynamics," *IEEE Transactions on Pattern Analysis and Machine Intelligence*, vol. 34, no. 1, pp. 33-50, 2012.

[8] S. H. Khan, X. He, F. Porikli, and M. Bennamoun, "Forest change detection in incomplete satellite images with deep neural networks," *IEEE Transactions on Geoscience and Remote Sensing*, vol. 55, no. 9, pp. 5407-5423, 2017.

[9] A. Shafique, G. Cao, Z. Khan, M. Asad, and M. Aslam, "Deep learning-based change detection in remote sensing images: A review," *Remote Sensing*, vol. 14, no. 4, 2022.

[10] Z. Lv, H. Huang, X. Li, M. Zhao, J. A. Benediktsson, W. Sun, and N. Falco, "Land cover change detection with heterogeneous remote sensing images: Review, progress, and perspective," *Proceedings of the IEEE*, vol. 110, no. 12, pp. 1976-1991, 2022.

[11] H. Chen, N. Yokoya, C. Wu, and B. Du, "Unsupervised multimodal change detection based on structural relationship graph representation learning," *IEEE Transactions on Geoscience and Remote Sensing*, vol. 60, pp. 1-18, 2022.

[12] Y. Sun, L. Lei, D. Guan, M. Li, and G. Kuang, "Sparse-constrained adaptive structure consistency-based unsupervised image regression for heterogeneous remote-sensing change detection," *IEEE Transactions on Geoscience and Remote Sensing*, vol. 60, pp. 1-14, 2022.

[13] Y. Sun, L. Lei, D. Guan, and G. Kuang, "Iterative robust graph for unsupervised change detection of heterogeneous remote sensing images," *IEEE Transactions on Image Processing*, vol. 30, pp. 6277-6291, 2021.

[14] P. J. Howarth and G. M. Wickware, "Procedures for change detection using Landsat digital data," *International Journal of Remote Sensing*, vol. 2, no. 3, pp. 277-291, 1981.

[15] R. D. Jackson, "Spectral indices in N-Space," *Remote Sensing of Environment*, vol. 13, no. 5, pp. 409-421, 1983.

[16] W. J. Todd, "Urban and regional land use change detected by using Landsat data," *Journal of Research of the U.S. Geological Survey*, vol. 5, no. 5, pp. 529-534, 1977.

[17] L. Wan, Y. Xiang, and H. You, "A post-classification comparison method for SAR and optical images change detection," *IEEE Geoscience and Remote Sensing Letters*, vol. 16, no. 7, pp. 1026-1030, 2019.

[18] D. Brunner, G. Lemoine, and L. Bruzzone, "Earthquake damage assessment of buildings using VHR optical and SAR imagery," *IEEE Transactions on Geoscience and Remote Sensing*, vol. 48, no. 5, pp. 2403-2420, 2010.

[19] Z. Liu, G. Li, G. Mercier, Y. He, and Q. Pan, "Change detection in heterogenous remote sensing images via homogeneous pixel transformation," *IEEE Transactions on Image Processing*, vol. 27, no. 4, pp. 1822-1834, 2018.

[20] J. Liu, M. Gong, K. Qin, and P. Zhang, "A deep convolutional coupling network for change detection based on heterogeneous optical and radar images," *IEEE Transactions on Neural Networks and Learning Systems*, vol. 29, no. 3, pp. 545-559, 2018.

[21] X. Jiang, G. Li, Y. Liu, X.-P. Zhang, and Y. He, "Change detection in heterogeneous optical and SAR remote sensing images via deep homogeneous feature fusion," *IEEE Journal of Selected Topics in Applied Earth Observations and Remote Sensing*, vol. 13, pp. 1551-1566, 2020.

[22] X. Li, Z. Du, Y. Huang, and Z. Tan, "A deep translation (GAN) based change detection network for optical and SAR remote sensing images," *ISPRS Journal of Photogrammetry and Remote Sensing*, vol. 179, pp. 14-34, 2021.

[23] Z.-G. Liu, Z.-W. Zhang, Q. Pan, and L.-B. Ning, "Unsupervised change detection from heterogeneous data based on image translation," *IEEE Transactions on Geoscience and Remote Sensing*, vol. 60, pp. 1-13, 2021.

[24] X. Niu, M. Gong, T. Zhan, and Y. Yang, "A conditional adversarial network for change detection in heterogeneous images," *IEEE Geoscience and Remote Sensing Letters*, vol. 16, no. 1, pp. 45-49, 2018.

[25] L. T. Luppino, M. Kampffmeyer, F. M. Bianchi, G. Moser, S. B. Serpico, R. Jenssen, and S. N. Anfinsen, "Deep image translation with an affinity-based change prior for unsupervised multimodal change detection," *IEEE Transactions on Geoscience and Remote Sensing*, vol. 60, pp. 1-22, 2021.

[26] L. T. Luppino, M. A. Hansen, M. Kampffmeyer, F. M. Bianchi, G. Moser, R. Jenssen, and S. N. Anfinsen, "Code-aligned autoencoders for unsupervised change detection in multimodal remote sensing images," *IEEE Transactions on Neural Networks and Learning Systems*, 2022.

[27] T. Schmidt, "Coping with copulas," *Copulas-From theory to application in finance*, Risk Books, London, pp. 3-34, 2007.

[28] G. Mercier, G. Moser, and S. B. Serpico, "Conditional copulas for change detection in heterogeneous remote sensing images," *IEEE Transactions on Geoscience and Remote Sensing*, vol. 46, no. 5, pp. 1428-1441, 2008.

[29] U. Cherubini, E. Luciano, and W. Vecchiato, *Copula methods in finance*. John Wiley & Sons, 2004.

[30] T. Janke, M. Ghanmi, and F. Steinke, "Implicit generative copulas," *Advances in Neural Information Processing Systems*, vol. 34, pp. 26 028- 26 039, 2021.

[31] Y. Ng, A. Hasan, K. Elkhalil, and V. Tarokh, "Generative archimedean copulas," in *Proceedings of the Thirty-Seventh Conference on Uncertainty in Artificial Intelligence*. PMLR, 2021, pp. 643-653.

[32] Z. Zeng and T. Wang, "Neural copula: A unified framework for estimating generic high-dimensional copula functions," *arXiv preprint arXiv:2205.15031*, 2022.

[33] C. Li, G. Li, X. Wang, and P. K. Varshney, "A copula-based method for change detection with multisensor optical remote sensing images," *IEEE Transactions on Geoscience and Remote Sensing*, vol. 61, pp. 1- 15, 2023.

[34] M. Sklar, "Fonctions de répartition à N dimensions et leurs marges," ` *Publ. inst. statist. univ. Paris*, vol. 8, pp. 229-231, 1959.

[35] S. Zhang, L. N. Theagarajan, S. Choi, and P. K. Varshney, "Fusion of correlated decisions using regular vine copulas," *IEEE Transactions on Signal Processing*, vol. 67, no. 8, pp. 2066-2079, 2019.

[36] S. G. Iyengar, P. K. Varshney, and T. Damarla, "A parametric copula-based framework for hypothesis testing using heterogeneous data," *IEEE Transactions on Signal Processing*, vol. 59, no. 5, pp. 2308-2319, 2011.

[37] X. Wang, D. Zhu, G. Li, X.-P. Zhang, and Y. He, "Proposal-copula-based fusion of spaceborne and airborne SAR images for ship target detection," *Information Fusion*, vol. 77, pp. 247-260, 2022.

[38] C. Rudin, C. Chen, Z. Chen, H. Huang, L. Semenova, and C. Zhong, "Interpretable machine learning: Fundamental principles and 10 grand challenges," *Statistic Surveys*, vol. 16, pp. 1-85, 2022.

[39] C. Xu, Z. Liao, C. Li, X. Zhou, and R. Xie, "Review on interpretable machine learning in smart grid," *Energies*, vol. 15, no. 12, p. 4427, 2022.

[40] Y. Zhang, P. Tiňo, A. Leonardis and K. Tang, "A survey on neural network interpretability," *IEEE Transactions on Emerging Topics in Computational Intelligence*, vol. 5, no. 5, pp. 726–742, 2021.

[41] D. V. Carvalho, E. M. Pereira, and J. S. Cardoso, "Machine learning interpretability: A survey on methods and metrics," *Electronics*, vol. 8, no. 8, p. 832, 2019.

[42] M. Raissi, P. Perdikaris, and G. E. Karniadakis, "Physics-informed neural networks: A deep learning framework for solving forward and inverse problems involving nonlinear partial differential equations," *Journal of Computational Physics*, vol. 378, pp. 686-707, 2019.

[43] R. R. Selvaraju, M. Cogswell, A. Das, R. Vedantam, D. Parikh, and D. Batra, "Grad-CAM: Visual explanations from deep networks via gradient-based localization," in *2017 IEEE International Conference on Computer Vision (ICCV)*, 2017, pp. 618-626.

[44] A. Guo, R. Dian, and S. Li, "A deep framework for hyperspectral image fusion between different satellites," *IEEE Transactions on Pattern Analysis and Machine Intelligence*, vol. 45, no. 7, pp. 7939-7954, 2023.

[45] Z. Kuang, H. Bi, F. Li, C. Xu, and J. Sun, "Polarimetry-inspired contrastive learning for class-imbalanced PolSAR image classification," *IEEE Transactions on Geoscience and Remote Sensing*, vol. 62, pp. 1-19, 2024.

[46] L. Mou, P. Ghamisi, and X. X. Zhu, "Deep recurrent neural networks for hyperspectral image classification," *IEEE Transactions on Geoscience and Remote Sensing*, vol. 55, no. 7, pp. 3639-3655, 2017.

[47] Y. Pang, J. Lin, T. Qin, and Z. Chen, "Image-to-image translation: Methods and applications," *IEEE Transactions on Multimedia*, vol. 24, pp. 3859-3881, 2022.

[48] J. Y. Zhu, T. Park, P. Isola, and A. A. Efros, "Unpaired image-to-image translation using cycle-consistent adversarial networks," in *2017 IEEE International Conference on Computer Vision (ICCV)*, 2017, pp. 2242-2251.

[49] J. C. Bezdek, R. Ehrlich, and W. Full, "FCM: The fuzzy *c*-means clustering algorithm," *Computers & geosciences*, vol. 10, no. 2-3, pp. 191-203, 1984.

[50] R. Achanta, A. Shaji, K. Smith, A. Lucchi, P. Fua, and S. Susstrunk, ¨ "SLIC superpixels compared to state-of-the-art superpixel methods," *IEEE Transactions on Pattern Analysis and Machine Intelligence*, vol. 34, no. 11, pp. 2274-2282, 2012.

[51] Y. Sun, L. Lei, D. Guan, G. Kuang, and L. Liu, "Graph signal processing for heterogeneous change detection," *IEEE Transactions on Geoscience and Remote Sensing*, vol. 60, pp. 1-23, 2022.

[52] S. Liu and M. E. Hodgson, "Satellite image collection modeling for large area hazard emergency response," *ISPRS Journal of Photogrammetry and Remote Sensing*, vol. 118, pp. 13-21, 2016.

[53] O. Ozdemir, T. G. Allen, S. Choi, T. Wimalajeewa, and P. K. Varshney, "Copula based classifier fusion under statistical dependence," *IEEE Transactions on Pattern Analysis and Machine Intelligence*, vol. 40, no. 11, pp. 2740-2748, 2018.

[54] B. W. Silverman, *Density estimation for statistics and data analysis*. Routledge, 2018.

[55] A. Krizhevsky, I. Sutskever, and G. E. Hinton, "ImageNet classification with deep convolutional neural networks," *Communications of the ACM*, vol. 60, no. 6, pp. 84-90, 2017.

[56] H. Abdi and L. J. Williams, "Principal component analysis," *WIREs Computational Statistics*, vol. 2, no. 4, pp. 433-459, 2010.

[57] R. Touati, M. Mignotte, and M. Dahmane, "Multimodal change detection in remote sensing images using an unsupervised pixel pairwise-based Markov random field model," *IEEE Transactions on Image Processing*, vol. 29, pp. 757-767, 2019.

[58] A. Gulli and S. Pal, *Deep learning with Keras*. Packt Publishing Ltd, 2017.

[59] C. Genest and J.-C. Boies, "Detecting dependence with kendall plots," *The American Statistician*, vol. 57, no. 4, pp. 275–284, 2003.